\documentclass{jlcl}

\newcommand{\authornames}{Simon Münker}

\newcommand{\authorlastnames}{Münker}

\newcommand{\articletitle}{Political Bias in LLMs: Unaligned Moral Values in Agent-centric Simulations}

\newcommand{\shortarticletitle}{Political Bias in LLMs}

\usepackage[utf8]{inputenc}
\usepackage[T1]{fontenc}
\usepackage[german, main=english]{babel} 
\usepackage{booktabs}
\usepackage[hidelinks]{hyperref}
\usepackage{orcidlink}

\usepackage[natbibapa]{apacite}

\usepackage{adjustbox}

\usepackage{graphicx}
\usepackage{booktabs}

\usepackage{tabularx}
\usepackage{multirow}
\usepackage{makecell}

\usepackage{enumitem}
\usepackage{float}

\newcommand{\jlclvolume}{}
\newcommand{\jlclnumber}{38(2)}
\newcommand{\jlclyear}{2025}

\begin{document}

\setcounter{page}{125}
\thispagestyle{plain}

\authordata


\noindent{\scriptsize
\textbf{This work was published in Journal for Language Technology and Computational Linguistics,} 2025, available online at \url{https://jlcl.org/article/view/289}. Please cite as: Münker, S. (2025). Political Bias in LLMs: Unaligned Moral Values in Agent-centric Simulations. Journal for Language Technology and Computational Linguistics, 38(2), 125–138. \url{https://doi.org/10.21248/jlcl.38.2025.289}
}

\section*{Abstract}
Contemporary research in social sciences increasingly utilizes state-of-the-art generative language models to annotate or generate content. While these models achieve benchmark-leading performance on common language tasks, their application to novel out-of-domain tasks remains insufficiently explored. To address this gap, we investigate how personalized language models align with human responses on the Moral Foundation Theory Questionnaire. We adapt open-source generative language models to different political personas and repeatedly survey these models to generate synthetic data sets where model-persona combinations define our sub-populations. Our analysis reveals that models produce inconsistent results across multiple repetitions, yielding high response variance. Furthermore, the alignment between synthetic data and corresponding human data from psychological studies shows a weak correlation, with conservative persona-prompted models particularly failing to align with actual conservative populations. These results suggest that language models struggle to coherently represent ideologies through in-context prompting due to their alignment process. Thus, using language models to simulate social interactions requires measurable improvements in in-context optimization or parameter manipulation to align with psychological and sociological stereotypes properly.
\section{Introduction}
Large Language Models (LLMs) have not only transformed consumer markets \citep{teubner2023welcome} but have also become influential tools within academic research where text serves as the primary subject of investigation \citep{tiunova2023chatgpt}. These systems demonstrate remarkable capabilities from classification and information extraction from unstructured data \citep{xu2023nir} to sophisticated text generation adaptable to various stylistic requirements \citep{bhandarkar2024emulating}. In social science research, a growing interest has emerged in utilizing LLMs to generate content that simulates specific user behaviors, particularly those associated with different political ideologies. A prevalent approach in this domain involves providing LLMs with abstract textual descriptions of political ideologies to guide their responses \citep{argyle2023out}. This method assumes that models can effectively generalize from these abstract descriptions to produce appropriate responses for tasks such as simulating social media content. However, current research lacks rigorous empirical verification of how consistently persona-based prompting can accurately represent individuals with specified ideological orientations. The fundamental assumption—that LLMs inherently encode ideological perspectives within their parameters—remains largely untested.

In contrast to computational approaches for assessing political ideology, differential psychology offers established frameworks for measuring human political orientations through abstract values and beliefs. Moral Foundation Theory (MFT) provides one such framework, measuring individuals' reliance on five distinct moral foundations \citep{graham2009liberals}. These foundations represent different sets of moral concerns that influence attitudes toward social and political issues. When combined with self-reported ideological identification, MFT demonstrates significant correlations between moral foundations and political orientation \citep{hatemi2019ideology}. If LLMs are to serve as effective proxies for human users, they should demonstrate consistent responses to standardized assessments like MFT questionnaires, aligning with patterns observed in human populations of corresponding ideological orientations.

The deployment of LLMs as human substitutes appears advantageous for studying online social networks (OSNs), as researchers can design controlled, text-centric environments for experimentation \citep{argyle2023out}. This approach offers a potential solution to challenges created by OSN providers' increasing restrictions on data access, which have hindered researchers' ability to conduct data-driven experiments using authentic user data \citep{bruns2021after}. However, we argue that uncritical application of market-driven technologies poses significant risks to research validity. Critical analysis of these models' performance in novel, out-of-domain tasks is essential before deploying them as simulated users in more complex applications. Without such foundational assessment, experiments utilizing synthetic OSN users provide limited insight into how accurately they represent genuine human interaction patterns.

\paragraph{Research Questions \& Contributions}
Our work establishes a foundation for analyzing how persona prompt modifications affect LLMs' representation of political ideologies across the left-right spectrum. We consider analyzing it a prerequisite to determining whether LLMs can effectively generalize from abstract ideological descriptions to specific applications, such as generating ideologically-consistent content or reactions. Our investigation focuses on two research questions:

\begin{description}
    \item[RQ$_1$] 
    How consistently do LLMs perform in their factory settings when surveyed with/without personas by only manipulating them through in-context prompting?

    \item[RQ$_2$]
    How closely do LLMs align in their factory settings by only manipulating them through in-context prompting to human participant groups?
    
\end{description}

Through systematic investigation of these questions, we contribute: (1) a methodological framework for evaluating LLMs as ideological simulacra using established psychological instruments; (2) empirical evidence regarding the consistency and human-alignment of different models across political personas; and (3) critical insights into the limitations of using persona-based prompting to represent complex ideological perspectives.
\section{Background}
We aim to connect our work to the existing critique of LLMs, with a focus on their application and the perception of their capabilities in terms of language understanding and ability to communicate. Further, we outline the unreflected application of synthetic users in the social sciences as human replacements and critique the expressiveness of those studies. 

\paragraph{Not more than stochastic parrots?}
\citet{bender2021dangers} critiqued that language models only manipulated textual content statistically to generate responses that give the impression of language understanding, like a parrot that listens to a myriad of conversations and anticipates how to react accordingly. Current conversational models are published by commercial facilities, with a business model relying on the illusion of models capable of language understanding and human-like conversation skills \citep{kanbach2024genai}. Thus, we have two extreme standpoints towards LLMs: a reductionist perspective that considers these models as next-word prediction machines based on matrix multiplication, and an anthropomorphic view that attributes human-like qualities to those systems \citep{bubeck2023sparks}. While we disagree with a (naive) anthropomorphism and current research questions the language understanding capabilities \citep{dziri2023faith}, we argue that when utilizing LLMs as human simulacra \citep{shanahan2024simulacra}, we must assume human-like qualities to a certain degree. Without this assumption, utilizing LLM agents to model interpersonal communication can only yield a shallow copy, a conversation between parroting entities.

\paragraph{LLMs as synthetic characters}
The usage of LLMs as human simulacra (representation) began with the application as non-player characters (NPCs) in a Sims-style game world to simulate interpersonal communication and day-to-day lives \citep{park2023generative}. The application of LLMs as synthetic characters has expanded beyond gaming environments into various fields of social science research \citep{argyle2023out}. Those disciplines already started to use these models as a replacement in social studies, arguing that conditioning through prompting causes the systems to accurately emulate response distributions from a variety of human subgroups \citep{argyle2023out}. While these applications show promise, they also raise significant methodological and ethical questions. Current research raises concerns about potential biases in the training data leading to misrepresentation of certain groups or viewpoints \citep{abid2021persistent, hutchinson2020social}. Without a deeper understanding of the model's representations of ideologies, we risk oversimplifying complex human behaviors and social dynamics. Especially as these approaches \citep{argyle2023out} ignore that LLMs lack embodiment in the physical world. This disembodied nature means they lack the grounding in physical reality – expressed by cultural contexts, physical environments, and interpersonal relationships – that shapes human cognition, perception, and decision-making \citep{hussein2012sapir}.

\section{Methods}
We repeatedly prompt LLMs to answer an MFT questionnaire with a neutral – model default – baseline and three different political persona system prompts to nudge the model toward a left-right ideology. Thus, we obtain a population for each model ($12$)/persona ($4$) combination that is the base for our variance and cross-human analysis. The populations contain $50$ samples. In total, we obtain $2,400$ artificially filled surveys.

\paragraph{Models}
Our research focuses on models with openly available weights that researchers can deploy locally using moderate computational infrastructure — specifically, systems with approximately $80$GB of video memory. These restrictions make our results and experiment pipeline usable for smaller research facilities without access to third-party providers. To broaden the selection across the size of models and their architecture, we include LLMs ranging from $7$B up to $176$B parameters and include models based on a mixture of expert architecture \citep{du2022glam}. While commercial models like ChatGPT or Claude could provide valuable comparison points, we explicitly focus on open-weight models to ensure reproducibility and avoid dependency on potentially changing API behaviors or undisclosed model updates.

\paragraph{Questionnaire}
The center of our experiments forms the Moral Foundations Questionnaire (MFQ) originally proposed by \citep{graham2009liberals}. We attach the full version in appendix \ref{sec:app:questionnaire}. The MFQ is a psychological assessment tool designed to measure the degree to which individuals rely on five different moral foundations when making moral judgments: care/harm (kindness, gentleness, nurturance), fairness/cheating (justice, rights, autonomy), loyalty/betrayal (solidarity, patriotism, sacrifice), authority/subversion (leadership, fellowship, authority), purity/degradation (living in a noble way). The questionnaire consists of 32 items divided into two parts. Moral Relevance: 16 questions asking participants to rate how relevant certain considerations are when making moral judgments. Moral Judgments: 16 questions asking participants to indicate their agreement or disagreement with specific moral statements. Responses are given on a 6-point Likert scale, ranging from 0 to 5. The Moral Relevance scale ranges from "not at all relevant" to "extremely relevant". By using a standardized and well-validated tool like the MFQ, we aim to provide a robust framework for comparing the moral reasoning capabilities of LLMs to those of human participants, while also exploring how different prompting strategies might influence these capabilities.

\paragraph{Human Baseline Data}
For our cross-human analysis, we utilize a dataset collected by \cite{graham2009liberals}, containing the aggregated responses from $1,613$ anonymous participants, a dataset collected by \cite{graham2011mapping}, containing aggregated responses $7,226$ from US-Americans and a dataset collected by \cite{kim2012moral}, containing $478$ South Koreans participants. All participants self-identified their political orientation on a spectrum from "very liberal" to "very conservative." We adopt the aggregation provided by the sources into liberal, moderate, and conservative groups to align with our model persona categories. The inclusion of South Korean participants allows for potential cross-cultural comparisons, though our primary analysis focuses on alignment with Western ideological categorizations that dominate most LLM training data.

\paragraph{Prompting}
The intention of our work is to assess synthetic surveys and evaluate the alignment between participants and language models. Thus, we opt for a simple prompt, containing only the task and an optional persona stating the political and ethical ideology. With the reduction to the keywords of the political ideology, we force the system to tap into its built-in concepts without modifying them in context \citep{wei2021finetuned}. The persona description prompts the model to represent the opinion of a \textit{"politically and ethically \{Conservative\;|\;Moderate\;|\;Liberal\}"}. We prompt the model on each question individually paired with the task description. For the base condition, we omit this description and simply ask the model to answer the question on the specified scale.

\subparagraph{Example-Prompt} \textit{\small
    You are a politically and ethically Conservative.\\
    Label the sentence with whether you agree or disagree. Choose from the following labels: [0] strongly disagree, [1] moderately disagree, [2] slightly disagree, [3] slightly agree, [4] moderately agree, [5] strongly agree. Respond only with the predicted class (1, 2, 3, 4, 5).\\
    Question: Compassion for those who are suffering is the most crucial virtue.
}

\section{Results}
Our response variance results (Table \ref{tab:questionnaire:moral-foundations:var:model:persona}) show a significant difference between the different models and personas. While Mistral 8x7B shows the highest stability with the lowest variance (0.030), Qwen 72B has a 14 times higher (0.425) variance. Also, adding ideological personas consistently increased response variance, with moderate personas (0.372) showing the most significant deviation from baseline responses (0.150). This higher variance for the moderate persona might reflect the ambiguity inherent in the term "moderate" across political contexts, as opposed to the more polarized liberal and conservative labels. For context, the variance values in Table \ref{tab:questionnaire:moral-foundations:var:model:persona} represent how consistently each model-persona combination answered the same questions across multiple trials. Lower variance indicates more stable and predictable responses, which would be expected if the models had a coherent understanding of the political ideology they were prompted to represent.

Table \ref{tab:questionnaire:moral-foundations:cross-evaluation} presents the comparison between our model-generated responses and the human baseline data from \cite{graham2009liberals, graham2011mapping, kim2012moral}. The values represent the mean squared error between model responses and corresponding human population responses across the five moral foundations. Lower values indicate better alignment. The cross-evaluation shows that on average the models exhibited left-leaning bias, the mean liberal human to liberal model distance is 0.665 and the mean conservative distance is 0.972 – as reported for the GPT-family \citep{mcgee2023chat, rutinowski2024self}. Notably, our results show limited alignment with South Korean participants across all model-persona combinations (0.859) in contrast to US citizens (0.808), suggesting either cultural limitations in the models' training data or potentially different interpretations of political identity terms across cultures. Across all model sizes (7B to 176B parameters), we found no consistent correlation between model size and either response consistency or alignment with human baseline data. This finding challenges the common assumption that larger models necessarily perform better on tasks requiring nuanced understanding of human values and beliefs.

\begin{table}
    \centering
    \scriptsize
    \begin{tabular}{l | rrrr | r}
        \toprule
        persona       & base      & conservative & liberal   & moderate  & MEAN      \\
        model         &           &              &           &           &           \\
        \midrule
        Gemma 7B      & 0.073     & 0.134        & 0.061     & 0.057     & 0.081     \\
        Llama2 70B    & 0.309     & 0.514        & 0.422     & 0.447     & 0.423     \\
        Llama3 70B    & 0.116     & 0.062        & 0.089     & 0.300     & 0.141     \\
        Mistral 7B    & 0.259     & 0.665        & 0.204     & 0.489     & 0.404     \\
        Mixtral 8x22B & 0.162     & 0.134        & 0.112     & 0.180     & 0.147     \\
        Mixtral 8x7B  & 0.025     & 0.037        & 0.047     & 0.012     & 0.030     \\
        Qwen 72B      & 0.108     & 0.116        & 0.356     & 1.122     & 0.425     \\
        \midrule
        MEAN          & 0.150     & 0.237        & 0.184     & 0.372     & 0.236     \\
        \bottomrule
    \end{tabular}
    \caption{Response variance aggregated across questions by model and persona.}
    \label{tab:questionnaire:moral-foundations:var:model:persona}
\end{table}
\section{Discussion}
The inconsistency in model responses, particularly evident in Qwen, raises concerns about the reliability of using LLMs as proxies for human participants in social science research. Crucially, larger models did not consistently outperform smaller ones in our study. This finding challenges the common assumption that scaling model size leads to better performance in tasks requiring a nuanced understanding of human values and beliefs. Even the largest models in our study (up to 176B parameters) showed similar limitations in representing coherent political ideologies compared to much smaller alternatives. While our results show that Mixtral produces the most human-like and consistent responses across our model selection, the overall alignment between model outputs and human participant ideologies is limited. It highlights the restriction of prompting approaches to align LLMs with complex human belief systems and indicates that these systems do not have a built-in concept of those ideologies, at least not capturable using our proposed approach.

\paragraph{Political Biases}
Our results demonstrate a systematic pattern where models show a smaller average distance to liberal human groups than to conservative groups across all model-persona combinations, as shown in Table \ref{tab:questionnaire:moral-foundations:cross-evaluation}. This aligns with previous findings that commercial models like ChatGPT exhibit left-leaning tendencies \citep{mcgee2023chat, rutinowski2024self}. Such bias could lead to over-representation of progressive viewpoints in applications where these models generate content intended to represent diverse ideological perspectives. In simulated social network environments, this bias might affect not only the content these models generate but potentially the way they would process and respond to ideologically diverse inputs if used to simulate interactions between different political viewpoints. The imbalance in representation of political orientations might stem from the distribution of ideological content in training corpora, where progressive perspectives may be more prevalent or systematically favored during alignment processes.

\paragraph{Cultural Limitations}
The inclusion of South Korean participants in our cross-evaluation revealed consistently poorer alignment between model-generated responses and this population across all model-persona combinations. This suggests that the models may have an implicit Western bias in their understanding of political identities and moral foundations. Such cultural limitations are particularly problematic when considering the global application of LLM-based research and highlight the need for more diverse training data and evaluation metrics.

\paragraph{RQ$_1$}
LLMs showed varying levels of consistency in their performance when surveyed with and without personas through in-context prompting. The base (no persona) condition showed the lowest average variance, while adding personas increased response variance significantly, with moderate personas showing the highest average variance. These findings suggest that LLMs' consistency can be significantly affected by incorporating textual personas through prompting, and this effect varies considerably across different models. The observed variation could be due to biases in training data, limitations in model architecture, or fundamental challenges in representing complex moral concepts computationally.

\paragraph{RQ$_2$}
While Mixtral models showed the best overall alignment, there is no clear, consistent pattern of specific model-persona combinations aligning well with particular human participant groups. This suggests that simple prompt-based persona modifications may not be sufficient to accurately represent diverse human ideologies and moral foundations. The observed misalignment between model outputs and human responses may be partially attributed to representational limitations in LLMs. These models, trained on human-generated data, may inadvertently reflect and amplify certain patterns in the data without necessarily developing coherent computational representations of complex ideological frameworks. Based on our observations, we can hardly justify using in-context prompted language models to simulate human ideologies without further research. Previous work on human simulacra \citep{argyle2023out, park2023generative} investigates the generated content or opinions on a superficial level but omits questioning whether LLMs can accurately represent the underlying belief systems and thought processes that characterize different ideological positions.

\paragraph{Variance: The lower the better?}
The preceding results and discussion focus on the observed variance in the collected data. Our analysis generally assumes a lower variance as the favorable outcome, indicating a more robust and consistent representation of the given ideology when answering the questionnaire. However, when considering LLMs as human simulacra, this reliability may not always be desirable. Human responses to moral questions naturally contain some variance, both within individuals over time and between individuals who identify with the same political ideology. Future research should establish benchmarks for "human-like" levels of response variance to better evaluate whether LLMs' inconsistency represents a limitation or potentially a more realistic simulation of human cognitive processes. This represents an important direction for follow-up studies that could compare the variance patterns in human populations to those observed in our model populations.

\paragraph{Ethical Considerations}
The use of LLMs to impersonate political personas raises several specific ethical concerns that researchers and developers should address. First, the potential for misrepresentation of ideological groups could reinforce stereotypes or create caricatures rather than authentic representations of diverse viewpoints. Second, as these technologies become more widespread, they could be misused to artificially inflate apparent consensus around certain political positions by generating large volumes of seemingly diverse but actually biased content. Third, the observed Western bias in ideological representation risks marginalizing non-Western perspectives in global discourse. Finally, there are privacy and consent issues around using models to simulate specific demographic groups who have not explicitly consented to such representation. Researchers employing LLMs as human simulacra must implement transparent documentation of model limitations and biases, establish clear guidelines for appropriate applications, and develop evaluation frameworks that assess ideological representation beyond surface-level content generation.

\paragraph{Conclusion}
Our results indicate that researchers must remain cautious and critical when applying these models in social science contexts, considering the ethical implications and potential limitations outlined above. Based on our findings, we argue that utilizing in-context prompted LLMs as human simulacra currently provides an inadequate representation of abstract political ideologies and human discourses, resulting in only a superficial simulation of genuine ideological diversity. Reducing interpersonal communication to computational models lacking embodied experience and trained primarily through statistical pattern recognition risks oversimplifying the complex nature of human moral and political reasoning. Importantly, our work demonstrates that this limitation persists regardless of model size, suggesting that simply scaling up parameters is unlikely to solve the fundamental challenges of representing human ideological perspectives without more sophisticated approaches to model development and evaluation.
\section*{Acknowledgments}
We thank Nils Schwager and Kai Kugler for our constructive discussions and Achim Rettinger for providing the research environment. This work is fully supported by TWON (project number 101095095), a research project funded by the European Union under the Horizon framework (HORIZON-CL2-2022-DEMOCRACY-01-07).

\section*{Limitations}
The scope of our findings is necessarily constrained by several methodological factors. First, our experiment includes only a subset of available open-source LLMs, and results may differ with other architectures or proprietary models. Second, our assessment of political alignment relies exclusively on the Moral Foundations Theory questionnaire, which, while validated in psychological research, represents only one framework for measuring political orientation. Alternative instruments might yield different insights or patterns of alignment. Third, our persona prompting technique employs minimal ideological descriptors, and more elaborate prompting strategies might produce different results. Additionally, our cross-cultural comparison was limited to Western and South Korean populations, potentially overlooking important cultural nuances in moral reasoning across other regions. Finally, the inherent limitations of LLMs—their lack of embodiment, experiential learning, and authentic human socialization—fundamentally restrict their ability to genuinely represent human moral and political reasoning processes.

\section*{Ethics Statement}
This research was conducted in accordance with the ACM Code of Ethics. The raw results, implementation details, and code-base are available upon request from the corresponding author (\href{mailto: muenker@uni-trier.de}{muenker@uni-trier.de}). We acknowledge the ethical complexities of using AI to simulate human political perspectives and have made efforts to interpret our findings with appropriate caution, avoiding overstatement of LLMs' capabilities to represent human belief systems. We emphasize that our work should not be used to justify the replacement of diverse human participants in social science research with AI-generated responses, as our findings specifically highlight the limitations of such approaches. Furthermore, we recognize the potential for misuse of persona-based LLM applications in political contexts and advocate for continued critical examination of these technologies as they evolve.


\appendix
\begin{table}[H]
    \section{Human \& LLM Cross-Evaluation}
    \centering
    \includegraphics[width=\textwidth]{./assets/heat.cross_evaluation.pdf}
    \caption{
        Absolute difference (lower is better) between the moral foundation scores of the selected models and scores across political ideologies of anonymous participants \citep{graham2009liberals}, US-Americans \citep{graham2011mapping} and Koreans \citep{kim2012moral}. The scale ranges between $0$ (no distance between model and human) and $5$ (maximum distance).
    }
    \label{tab:questionnaire:moral-foundations:cross-evaluation}
\end{table}

\newpage
\section{MFQ \citep{graham2009liberals}}
\label{sec:app:questionnaire}

{\footnotesize

\textbf{Agreement:}
For each of the statements below, please indicate how well each statement describes you or your opinions. Response options: Strongly disagree (0); Moderately disagree (1); Slightly disagree (2); Slightly agree (3); Moderately agree (4); and Strongly agree (5).

\begin{enumerate}[itemsep=0.0025cm, leftmargin=0.8cm]
    \item Compassion for those who are suffering is the most crucial virtue.
    \item When the government makes laws, the number one principle should be ensuring that everyone is treated fairly.
    \item I am proud of my country's history.
    \item Respect for authority is something all children need to learn.
    \item People should not do things that are disgusting, even if no one is harmed.
    \item It is better to do good than to do bad.
    \item One of the worst things a person could do is hurt a defenseless animal.
    \item Justice is the most important requirement for a society.
    \item People should be loyal to their family members, even when they have done something wrong.
    \item Men and women each have different roles to play in society.
    \item I would call some acts wrong on the grounds that they are unnatural.
    \item It can never be right to kill a human being.
    \item I think it's morally wrong that rich children inherit a lot of money while poor children inherit nothing.
    \item It is more important to be a team player than to express oneself.
    \item If I were a soldier and disagreed with my commanding officer's orders, I would obey anyway because that is my duty.
    \item Chastity is an important and valuable virtue.
\end{enumerate}

\noindent\textbf{Relevance:}
When you decide whether something is right or wrong, to what extent are the following considerations relevant to your thinking? Response options: Not at all relevant (0); Not very relevant (1); Slightly relevant (2); Somewhat relevant (3); Very relevant (4); and Extremely relevant (5).

\begin{enumerate}[itemsep=0.0025cm, leftmargin=0.8cm]
    \item[17.] Whether or not someone suffered emotionally.
    \item[18.] Whether or not some people were treated differently than others.
    \item[19.] Whether or not someone's action showed love for his or her country.
    \item[20.] Whether or not someone showed a lack of respect for authority.
    \item[21.] Whether or not someone violated standards of purity and decency.
    \item[22.] Whether or not someone was good at math.
    \item[23.] Whether or not someone cared for someone weak or vulnerable.
    \item[24.] Whether or not someone acted unfairly.
    \item[25.] Whether or not someone did something to betray his or her group.
    \item[26.] Whether or not someone conformed to the traditions of society.
    \item[27.] Whether or not someone did something disgusting.
    \item[28.] Whether or not someone was cruel.
    \item[29.] Whether or not someone was denied his or her rights.
    \item[30.] Whether or not someone showed a lack of loyalty.
    \item[31.] Whether or not an action caused chaos or disorder.
    \item[32.] Whether or not someone acted in a way that God would approve of.
\end{enumerate}

\noindent\textbf{Scoring:} 
Average each of the following items to get five scores corresponding with the five foundations, plus one catch score.

\vspace*{0.2cm}

\centering
\begin{tabular}{lll}
     \textbf{Harm:} 1, 7, 12, 17, 23, 28 & 
     \textbf{Ingroup:} 3, 9, 14, 19, 25, 30 & 
     \textbf{Purity:} 5, 11, 16, 21, 27, 32 \\
     \textbf{Fairness:} 2, 8, 13, 18, 24, 29 &  
     \textbf{Authority:} 4, 10, 15, 20, 26, 31 &
     \textbf{Catch:} 6, 22
\end{tabular}

}


\newpage
\bibliographystyle{apacite}
\bibliography{references}


\clearpage
\noindent \textbf{Correspondence}
\hspace{12pt}

\noindent Simon Münker \orcidlink{0000-0003-1850-5536}\\[0.5\baselineskip]
Trier University \\
Department of Computational Linguistics \\ 
Trier, Germany\\
\href{mailto: muenker@uni-trier.de}{muenker@uni-trier.de}


\end{document}